\documentclass[acmsmall]{acmart}

\usepackage[utf8]{inputenc} 
\usepackage[T1]{fontenc}    
\usepackage{hyperref}       
\usepackage{url}            
\usepackage{booktabs}       
\usepackage{amsfonts}       
\usepackage{nicefrac}       
\usepackage{microtype}      

\usepackage{url}

\usepackage{multirow}
\usepackage{longtable} 

\newcommand{\change}[1]{#1}

\title{Challenges in Deploying Machine Learning: a Survey of Case Studies}

\acmJournal{CSUR}

\author{Andrei Paleyes}
\affiliation{%
	\institution{University of Cambridge}
	\department{Department of Computer Science and Technology}
	\city{Cambridge}
	\country{United Kingdom}
}
\email{ap2169@cam.ac.uk}

\author{Raoul-Gabriel Urma}
\affiliation{%
	\institution{Cambridge Spark}
	\city{Cambridge}
	\country{United Kingdom}
}
\email{raoul@cambridgespark.com}

\author{Neil D. Lawrence}
\affiliation{%
	\institution{University of Cambridge}
	\department{Department of Computer Science and Technology}
	\city{Cambridge}
	\country{United Kingdom}
}
\email{ndl21@cam.ac.uk}

\setcopyright{rightsretained}
\acmJournal{CSUR}
\acmYear{2022} \acmVolume{1} \acmNumber{1} \acmArticle{1} \acmMonth{1} \acmPrice{}
\acmDOI{10.1145/3533378}

\ccsdesc[500]{Computing methodologies~Machine learning}
\keywords{Machine learning applications, sofware deployment}

\begin{document}

\begin{abstract}
In recent years, machine learning has transitioned from a field of academic research interest to a field capable of solving real-world business problems. However, the deployment of machine learning models in production systems can present a number of issues and concerns. This survey reviews published reports of deploying machine learning solutions in a variety of use cases, industries and applications and extracts practical considerations corresponding to stages of the machine learning deployment workflow. By mapping found challenges to the steps of the machine learning deployment workflow we show that practitioners face issues at each stage of the deployment process. The goal of this paper is to lay out a research agenda to explore approaches addressing these challenges.
\end{abstract}

\maketitle

\section{Introduction}
Machine learning (ML) has evolved from \change{being} an area of academic research to an applied field. According to a recent global survey conducted by McKinsey, machine learning is increasingly adopted in standard business processes with nearly 25 percent year-over-year growth \cite{cam2019global} and with growing interest from the general public, business leaders \cite{davenport2018artificial} and governments \cite{royal2017machine}.

This shift comes with challenges. Just as with any other field, there are significant differences between what works in an academic setting and what is required by a real world system. Certain bottlenecks and invalidated preconceptions should always be expected in the course of that process. As more solutions are developed and deployed, practitioners report their experience in various forms, including publications and blog posts. \change{Motivated by such reports and our personal experiences, in this study we aim to survey} the challenges in deploying machine learning in production\footnote{By \textit{production} we understand the setting where a product or a service are made available for use by its intended audience.} \change{with the objective of understanding what parts of the deployment process cause the most difficulties}. First, we provide an overview of the machine learning deployment workflow. Second, we review case studies to extract problems and concerns practitioners have at each particular deployment stage. Third, we discuss cross-cutting aspects that affect every stage of the deployment workflow: ethical considerations, law, end users' trust and security. Finally, we conclude with a brief discussion of potential solutions to these issues and further work.

Ours is not the first survey of machine learning in production. A decade ago\change{,} such surveys were already conducted, albeit with a different purpose in mind. Machine learning was mainly a research discipline, and it was uncommon to see ML solution\change{s} deployed for a business problem outside of ``Big Tech'' companies in the information technology industry. So\change{,} the purpose of such a survey was to show that ML can be used to solve a variety of problems, and illustrate it with examples, as was done by P{\v{e}}chou{\v{c}}ek and Ma{\v{r}}{\'\i}k \cite{pvechouvcek2008industrial}. Nowadays the focus has changed: machine learning is commonly adopted in many industries, and the question becomes not ``Where is it used?'', but rather ``How difficult is it to use?''

One approach to assess the current state of machine learning deployment for businesses is a survey conducted among professionals. Such surveys are primarily undertaken by private companies, and encompass a variety of topics. Algorithmia's report (\cite{wiggers2019algorithmia, hecht2019additup}) goes deep into \change{the} deployment timeline, with the majority of companies reporting between 8 and 90 days to deploy a single model, and 18\% taking even more time. A report by IDC \cite{wiggers2019idc} surveyed 2,473 organizations and found that a significant portion of their attempted AI deployments fail, quoting lack of expertise, bias in data and high costs as primary reasons. O'Reilly has conducted an interview study that focused on ML practitioners' work experience and tools they use \cite{lorica2018state}. Broader interview-based reports have also been produced by dotscience \cite{dotscience2019state} and dimensional research \cite{dimensional2019ai}.

While there are a number of business reports on the topic, the challenges of the entire machine learning deployment pipeline are not covered nearly as widely in the academic literature. There is a burgeoning field of publications focusing on specific aspects of deployed ML, such as Bhatt et al. \cite{bhatt2019explainable} which focuses on explainable ML or Amershi et al. \cite{amershi2019software} which discusses software engineering aspects of deploying ML. A large number of industry-specific surveys also exist, and we review some of these works in the appropriate sections below. However, the only general purpose survey we found is Baier et al. \cite{baier2019challenges}, which combines \change{a} literature review and interviews with industry partners. In contrast with that paper, which concentrates on \change{the} experience of \change{the} information technology sector, we aim at covering case studies from a wide variety of industries\change{, thus increasing the breadth of our survey}. We also discuss the most commonly reported challenges in the literature in much greater detail.

In our survey we consider three main types of papers:
\begin{itemize}
	\item Case study papers that report experience from a single ML deployment project. Such works usually go deep into discussing each challenge the authors faced and how it was overcome.
	\item Review papers that describe applications of ML in a particular field or industry. These reviews normally give a summary of challenges that are most commonly encountered during the deployment of the ML solutions in the reviewed field.
	\item ``Lessons learned'' papers where authors reflect on their past experiences of deploying ML in production.
\end{itemize}
To ensure that this survey focuses on the current challenges, only papers published in the last five years are considered, with only \change{a} few exceptions to this rule. We also refer \change{to} other types of papers where it is appropriate, e.g. practical guidance reports, interview studies and regulations. We have not conducted any interviews ourselves as part of this study. Further work and extensions are discussed at the end of this paper.

\change{The} main contribution of our paper is to show that practitioners face challenges at each stage of the machine learning deployment workflow. This survey supports our goal to raise awareness in the academic community of the variety of problems that practitioners face when deploying machine learning and start a discussion on what can be done to address these problems.

\section{Machine Learning Deployment Workflow}
For the purposes of this work we are using the ML deployment workflow definition suggested by Ashmore et al. \cite{ashmore2021assuring}, however it is possible to conduct a similar review with any other ML pipeline description, such as CRISP-DM \cite{shearer2000crisp} or TDSP \cite{severtson2017team}. In this section we give a brief overview of the definition we are using.

According to Ashmore et. al. \cite{ashmore2021assuring}, the process of developing an ML-based solution in an industrial setting consists of four stages\footnote{There is a requirements formulation stage that Ashmore et al. do not consider a part of the ML workflow. Similarly, we see this as a challenge across many domains and thus out of scope of this study. We refer interested readers to the work by Takeuchi and Yamamoto \cite{takeuchi2020business}.}:
\begin{itemize}
	\item \textbf{Data management}, which focuses on preparing data that is needed to build a machine learning model;
	\item \textbf{Model learning}, where model selection and training happens;
	\item \textbf{Model verification}, the main goal of which is to ensure \change{the} model adheres to certain functional and performance requirements;
	\item \textbf{Model deployment}, which is about integration of the trained model into the software infrastructure that is necessary to run it. This stage also covers questions around model maintenance and updates.
\end{itemize}

Each of these stages is broken down further into smaller steps. It is important to highlight that the apparent sequence of this description is not necessarily the norm in a real-life scenario. It is perfectly normal for these stages to run in parallel to a certain degree and inform each other via feedback loops. Therefore this or any similar breakdown should be considered not as a timeline, but rather as a useful abstraction to simplify references to concrete parts of the deployment pipeline. \change{We have chosen this representation of the workflow because comparing to alternatives it defines a larger number of deployment steps, which allows for finer classification of the challenges.}

For the remainder of this paper we discuss common issues practitioners face at each step. We also discuss cross-cutting aspects that can affect every stage of the deployment pipeline. Where appropriate, the concrete class of ML problems is specified (such as supervised learning or reinforcement learning), although \change{the} majority of the challenges can be encountered during the deployment of many types of learning tasks. Table~\ref{tab:summary} provides a summary of the issues and concerns we discuss.  By providing illustrative examples for each step of the workflow we show how troublesome the whole deployment experience can be. Note that the order in which challenges appear in that table and in the paper does not necessarily reflect their severity. \change{The} impact that a particular issue can have on a deployment project depends on a large variety of factors, such as business area, availability of resources, team experience, and more. For that reason we do not attempt to prioritize \change{the} challenges discussed in the survey.

\begin{longtable}{ |l|l|l| }
	\caption{All considerations, issues and concerns explored in this study. Each is assigned to the stage and step of the deployment workflow where it is commonly encountered.} \label{tab:summary} \\
	\hline
	\textbf{Deployment Stage} & \textbf{Deployment Step} & \textbf{Considerations, Issues and Concerns} \\
	\hline
	\endfirsthead
	\caption{Continued: All considerations, issues and concerns explored in this study. Each is assigned to the stage and step of the deployment workflow where it is commonly encountered.} \\
	\hline
	\textbf{Deployment Stage} & \textbf{Deployment Step} & \textbf{Considerations, Issues and Concerns} \\
	\hline
	\endhead
	\hline
	Data management & Data collection & Data discovery \\ 
	\cline{2-3}
	& Data preprocessing & Data dispersion \\
	& & Data cleaning \\
	\cline{2-3}
	& Data augmentation & Labeling of large volumes of data \\
	& & Access to experts \\
	& & Lack of high-variance data \\
	\cline{2-3}
	& Data analysis & Data profiling\\
	\hline
	Model learning & Model selection & Model complexity\\
	& & Resource-constrained environments \\ 
	& & Interpretability of the model \\
	\cline{2-3}
	& Training & Computational cost \\
	& & Environmental impact \\
	& & Privacy-aware training \\
	\cline{2-3}
	& Hyper-parameter selection & Resource-heavy techniques\\
	& & Unknown search space \\
	& & Hardware-aware optimization \\
	\hline
	Model verification & Requirement encoding & Performance metrics \\
	& & Business driven metrics \\
	\cline{2-3}
	& Formal verification & Regulatory frameworks \\
	\cline{2-3}
	& Test-based verification & Simulation-based testing \\
	& & Data validation routines \\
	& & Edge case testing \\
	\hline
	Model deployment & Integration & Operational support \\
	& & Reuse of code and models \\
	& & Software engineering anti-patterns \\
	& & Mixed team dynamics \\
	\cline{2-3}
	& Monitoring & Feedback loops \\ 
	& & Outlier detection \\
	& & Custom design tooling \\
	\cline{2-3}
	& Updating & Concept drift \\
	& & Continuous delivery \\
	\hline
	Cross-cutting aspects & Ethics & Aggravation of biases \\
	& & Fairness and accountability \\
	& & Authorship \\
	& & Decision making \\
	\cline{2-3}
	& Law & Country-level regulations \\
	& & Abiding by existing legislation \\
	& & Focus on technical solution only \\
	\cline{2-3}
	& End users' trust & Involvement of end users \\
	& & User experience \\
	& & Explainability score \\
	\cline{2-3}
	& Security & Data poisoning \\
	& & Model stealing \\
	& & Model inversion \\
	\hline
\end{longtable}

\section{Data Management}\label{sec:data}
Data is an integral part of any machine learning solution. \change{The} overall effectiveness of the solution depends on the training and test data as much as on the algorithm. The process of creating quality datasets is usually the very first stage in any production ML pipeline. Unsurprisingly, practitioners face a range of issues while working with data as reported by Polyzotis et al.~\cite{polyzotis2018data}. Consequently, this stage consumes time and energy that is often not anticipated beforehand. In this section, we describe issues concerning four steps within data management: data collection, data preprocessing, data augmentation and data analysis. Note that we consider storage infrastructure challenges, such as setting up databases and query engines, beyond the scope of this survey. We refer readers to the survey by Cai et al. \cite{cai2016iot} for further discussion of big data storage.

\subsection{Data collection}\label{sec:data-collection}
Data collection involves activities that aim to discover and understand what data is available, as well as how to organize convenient storage for it. The task of discovering what data exists and where it is can be a challenge by itself, especially in large production environments typically found within organizations \cite{delve-7}. Finding data sources and understanding their structure is a major task, which may prevent data scientists from even getting started on the actual application development. As explained by Lin and Ryaboy \cite{lin2013scaling}, at Twitter this situation often happened as a result of the same entity (e.g. a Twitter user) being processed by multiple services. Internally Twitter consists of multiple services calling each other, and every service is responsible for a single operation. This approach, known in software engineering as \change{the} ``single responsibility principle'' \cite{martin2002single}, results in an architecture that is very flexible in terms of scalability and modification. However, the flip side of this approach is that \change{at a large scale} it is very hard to keep track of what data related to the entity is being stored by which service, and in which form. Some data may only exist in a form of logs, which by their nature are not easily parsed or queried. An even worse case is the situation when data is not stored anywhere, and to build a dataset one needs to generate synthetic service API calls. Such \change{dispersion} of data creates major hurdles for data scientists, because without a clear idea of what data is available or can be obtained it is often impossible to understand what ML solutions can achieve.

\subsection{Data preprocessing}
The preprocessing step normally involves a range of data cleaning activities: identification of a schema, imputation of missing values, reduction of data into an ordered and simplified form, and mapping from raw form into a more convenient format. Methods for carrying out data manipulations like this is an area of research that goes beyond the scope of this study. We encourage readers to refer to review papers on the topic, such as Abedjan et al. \cite{abedjan2016detecting}, Patil and Kulkarni \cite{patil2012review}, Ridzuan et al. \cite{ridzuan2019review}.

A lesser known but important problem that can also be considered an object of the preprocessing step is data dispersion. It often turns out that there can be multiple relevant separate data sources \change{that} may have different schemas, different conventions, and their own way of storing and accessing the data. Joining this information into a single dataset suitable for machine learning can be a complicated task in its own right, known as the data integration process \cite{nazabal2020data}. An example of this is what developers of Firebird faced \cite{madaio2016firebird}. Firebird is an advisory system in the Atlanta Fire Department, that helps identify priority targets for fire inspections. As a first step towards developing Firebird data was collected from 12 datasets including \change{the} history of fire incidents, business licenses, households and more. These datasets were combined to give a single dataset covering all relevant aspects of each property monitored by the Fire Department. Authors particularly highlight data joining as a difficult problem. Given the fact that buildings can be identified through their geospatial location, each dataset contained spatial data specifying \change{the} building's address. Spatial information can be presented in different formats, and sometimes contains minor differences such as different spellings. All this needed to be cleaned and corrected. These corrections can be highly time consuming and proved to be so in the Firebird project.

\subsection{Data augmentation}\label{sec:augmentation}
There are multiple reasons why data might need to be augmented, and in practice one of the most problematic ones is the absence of labels. A label is a value that \change{the} ML model seeks to predict from the input data in a classic supervised learning setting. Real-world data is often unlabeled, thus labeling turns out to be a challenge in its own right. We discuss three possible factors for lack of labeled data: limited access to experts, absence of high-variance data, and sheer volume.

Label assignment is difficult in environments that tend to generate large volumes of data, such as network traffic analysis. To illustrate a scale of this volume, a single 1-GB/s Ethernet interface can deliver up to 1.5 million packets per second. Even with a huge downsampling rate this is still a significant number, and each sampled packet needs to be traced in order to be labeled. This problem is described by Pacheco et al. \cite{8543584}, which surveys applications of machine learning to network traffic classification, with tasks such as protocol identification or attack detection. There are two main ways of acquiring data in this domain, and both are complicated for labeling purposes:
\begin{itemize}
	\item Uncontrolled, collecting real traffic. This approach requires complex tracking flows belonging to a specific application. Due to this complexity very few works implement reliable ground truth assignment for real traffic.
	\item Controlled, emulating or generating traffic. \change{This approach is very sensitive to the choice of tooling and its ability to simulate the necessary traffic. Studies have shown} that existing tools for label assignment can introduce errors into collected ML datasets \change{of network traffic data}, going as high as almost 100\% for certain applications \cite{dusi2011quantifying}. Moreover, these tools' performance degrades severely for encrypted traffic.
\end{itemize}

Access to experts can be another bottleneck for collecting high-quality labels. It is particularly true for areas where expertise mandated by the labeling process is significant, such as medical image analysis \cite{budd2019survey}. Normally multiple experts are asked to label a set of images, and then these labels are aggregated to ensure quality. This is rarely feasible for big datasets due to experts' availability. A possible option here is to use noisy label oracles \cite{du2010active} or weak annotations \cite{peyre2017weakly}, however these approaches provide imprecise labels, which can \change{lead} to \change{a} loss in quality of the model \cite{ren2020denoising}. Such losses are unacceptable in the healthcare industry, where even the smallest deviation can cause  catastrophic results (this is known as The Final Percent challenge according to Budd et al. \cite{budd2019survey}).

Lack of access to high-variance data \change{(data that covers large parts of the feature space)} can be among the main challenges one faces when deploying machine learning solution\change{s from the lab environment to the real world}. Dulac-Arnold et al. \cite{dulacarnold2019challenges} explain that this is the case for Reinforcement Lea\change{r}ning (RL). RL is an area of ML that focuses on intelligent agents that learn by taking actions and receiving feedback from their environment. Agents make decisions on what action to take next to maximize the reward based on their previous experience. It is common practice in RL research to have access to separate environments for training and evaluation of an agent. However, in practice all data comes from the real system, and the agent can no longer have a separate exploration policy - this is simply unsafe. Therefore the data available becomes low-variance - \change{very little of the state space is covered}. While this approach ensures safety, it means that \change{the} agent is not trained to recognize an unsafe situation and make \change{the} right decision in it. A practical example of this issue can be seen in the area of autonomous vehicle control \cite{kuutti2019survey}. There simulations are often used for training, but complex interactions, such as friction, can be hard to model, and small variations in simulation may result in \change{the} agent not being transferable to the real world.

Budd et al. \cite{budd2019survey} show that interface design directly impacts \change{the} quality of applications built to collect annotations for unlabeled data. They discuss a range of projects that collected labels for medical images, all of which benefited from a well designed user interface. The authors conclude that \change{the} end user interface plays a large part in \change{the} overall success of the annotation applications\footnote{For further discussion of the role user interface can play in adoption of an ML system, see Section~\ref{sec:end-user}\change{.}}.

\subsection{Data analysis}
Data needs to be analyzed to uncover potential biases or unexpected distribution shifts in it. Availability of high quality tools is essential for conducting any kind of data analysis. One area that practitioners find particularly challenging in that regard is visualization for data profiling \cite{kandel2012enterprise}. Data profiling refers to all activities associated with troubleshooting data quality, such as missing values, inconsistent data types and verification of assumptions. Despite obvious relevance to the fields of databases and statistics, there are still too few tools that enable \change{the} efficient execution of these data mining tasks. The need for such tools becomes apparent considering that, according to the survey conducted by Microsoft \cite{kim2017data}, data scientists think data issues are the main reason to doubt \change{the} quality of the overall work.

\section{Model Learning}
Model learning is the stage of the deployment workflow that enjoys the most attention within the academic community. As an illustration of the scale of the field's growth, the number of submissions to NeurIPS, \change{a} primary conference on ML methods, has quadrupled in six years, going from 1678 submissions in 2014 to 6743 in 2019 \cite{charrez2019neurips}. Nevertheless, there are still plenty of practical considerations that affect the model learning stage. In this section, we discuss issues concerning three steps within model learning: model selection, training and hyper-parameter selection.

\subsection{Model selection}\label{sec:model-selection}
In many practical cases the selection of a model is decided by one key characteristic of a model: complexity. Despite areas such as deep learning and reinforcement learning gaining in popularity with the research community, in practice simpler models are often chosen. Such models include shallow neural network architectures, simple approaches based on Principal Component Analysis (PCA), decision trees and random forests.

Simple models can be used as a way to prove the concept of the proposed ML solution and get the end-to-end setup in place. This approach reduces the time to get a deployed solution, allows the collection of important feedback and also helps avoid overcomplicated designs. This was the case reported by Haldar et al. \cite{Haldar_2019}. In the process of applying machine learning to AirBnB search, the team started with a complex deep learning model. The team was quickly overwhelmed by its complexity and ended up consuming development cycles. After several failed deployment attempts the neural network architecture was drastically simplified: a single hidden layer NN with 32 fully connected ReLU activations. Even such a simple model had value, as it allowed the building of a whole pipeline of deploying ML models in \change{a} production setting, while providing reasonably good performance\footnote{We discuss more benefits of setting up the automated deployment pipeline in Section~\ref{sec:updating}.}. Over time the model evolved, with a second hidden layer being added, but it still remained fairly simple, never reaching the initially intended level of complexity.

Another advantage that less complex models can offer is their relatively modest hardware requirements. This becomes a key decision point in resource constrained environments, as shown by Wagstaff et al. \cite{10.1145/3292500.3330656}. They worked on deploying ML models to a range of scientific instruments onboard \change{the} Europa Clipper spacecraft. Spacecraft design is always a trade-off between the total weight, robustness and the number of scientific tools onboard. Therefore computational resources are scarce and their usage has to be as small as possible. These requirements naturally favor the models that are light on computational demands. The team behind Europa Clipper used machine learning for three anomaly detection tasks, some models took time series data as input and some models took images, and on all three occasions simple threshold or PCA based techniques were implemented. They were specifically chosen because of their robust performance and low demand \change{for} computational power.

A further example of a resource-constrained environment is wireless cellular networks, where energy, memory consumption and data transmission are very limited. Most advanced techniques, such as deep learning, are not considered yet for practical deployment, despite being able to handle high dimensional mobile network data \cite{challita2019machine}.

The ability to interpret the output of a model into understandable business domain terms often plays a critical role in model selection, and can even outweigh performance considerations. For that reason decision trees (DT) \cite{quinlan1986induction}, which can be considered a fairly basic ML algorithm, are widely used in practice. DT present\change{s} \change{its} branching logic in \change{a} transparent way that resembles human decision process, which allows for interpretation and auditing. Hansson et al. \cite{hansson2016machine} describe several cases in manufacturing that adopt DT because of their interpretability.

Banking is another industry where DT find\change{s} extensive use. As an illustrative example, it is used by Keramati et al. \cite{keramati2016developing} where the primary goal of the ML application is predicting churn in an interpretable way through if-then rules. While it is easy to imagine more complicated models learning the eventual input-output relationship for this specific problem, interpretability is key requirement here because of the need to identify the features of churners. The authors found DT to be the best model to fulfill this requirement.

Nevertheless, deep learning (DL) is commonly used for practical background tasks that require analysis \change{of} a large amount of previously acquired data. This notion is exemplified by the field of unmanned aerial vehicles (UAV) \cite{carrio2017review}. \change{Data collected by image sensors is the most common UAV data modality being exploited by DL, because of the sensors' low cost, low weight, and low power consumption.} DL offers excellent capabilities for processing and presentation of raw images acquired from sensors, but computational resource demands still remain the main blocker for deploying DL as an online processing instrument on board UAVs.

\subsection{Training}\label{sec:training}
Model training is the process of feeding the chosen model with \change{a} collected dataset in order to learn certain patterns or representations of the data. One of the biggest concerns with \change{the} model training stage is the economic cost associated with carrying out the training procedure due to the computational resources required. This is often true in the field of natural language processing (NLP), as illustrated by Sharir et al. \cite{sharir2020cost}. The authors observe that while the cost of individual floating-point  operations is decreasing, the overall cost of training NLP is only growing. They took one of the state-of-the-art models in the field, BERT \cite{devlin2018bert}, and found out that depending on the chosen model size full training procedure can cost anywhere between \$50k and \$1.6m in cloud computing resources, which is unaffordable for most research institutions and many companies. The authors observe that training dataset size, number of model parameters and number of operations used by the training procedure are all contributing towards the overall cost. Of particular importance here is the second factor: novel NLP models are already using billions of parameters, and this number is expected to increase further in the near future \cite{benaich2019state}.

A related concern is raised by Strubell et al. \cite{strubell2019energy} regarding the impact the training of ML models has on the environment. By consuming more and more computational resources, ML model training is driving up \change{energy} consumption and greenhouse gas emissions. According to the estimates provided in the paper, one full training cycle utilizing neural architecture search emits \change{an} amount of CO\textsubscript{2} comparable to what four average cars emit in their whole lifetime. The authors stress how important it is for researchers to be aware of such impact of model training, and argue that the community should give higher priority to computationally efficient hardware and algorithms. Similar concerns around environmental impact are voiced by Bender et al. \cite{bender2021dangers}.

As more businesses start using ML techniques on their users' data, concerns are being raised over \change{the} privacy of data and how well individuals\change{'} sensitive information is preserved over the course of the model training process \cite{papernot2018sok}. Illustrating the gravity of this concern, Shokri et al. constructed an attack for membership inference, that is to determine if a given input record was a part of the model's training dataset. They verified it on models trained on leading ML-as-a-service providers \cite{shokri2017membership}, achieving anywhere from 70\% up to 94\% accuracy of membership inference. Consequently companies have to consider privacy-aware approaches, which most likely come at a cost of the model accuracy. Navigating the trade-off between privacy and utility is considered an open challenge for practitioners dealing with sensitive data \cite{avent2020automatic}. Some of the ways this trade-off is resolved in practice are differential privacy that explicitly corrupts the data \cite{dwork2006differential}, homomorphic encryption that restricts the class of learning algorithm but allows for training on encrypted data \cite{gentry2009fully}, and federated learning that distributes training across personal devices to preserve privacy, but thereby constrains the algorithms that can be used for model fitting \cite{konevcny2015federated}.

\subsection{Hyper-parameter selection}\label{sec:hpo}
In addition to parameters that are learned during the training process, many ML models also require hyper-parameters. Examples of such hyper-parameters are \change{the} depth of a decision tree, \change{the} number of hidden layers in a neural network or \change{the} number of neighbors in k-Nearest Neighbors classifier. Hyper-parameter optimization (HPO) is the process of choosing the optimal setting of these hyper-parameters. Most HPO techniques involve multiple training cycles of the ML model. This is computationally challenging because in the worst case the size of the HPO task grows exponentially: each new hyper-parameter adds a new dimension to the search space. As discussed by Yang and Shami \cite{yang2020hyperparameter}, these considerations make HPO techniques very expensive and resource-heavy in practice, especially for applications of deep learning. Even approaches like Hyperband \cite{li2017hyperband} or Bayesian optimization \cite{snoek2012practical}, that are specifically designed to minimize the number of training cycles needed, are not yet able to deal with the high dimensional searches that emerge when many hyper-parameters are involved. Large datasets complicate matters by leading to long training times for each search.

Many hyper-parameter tuning approaches require the user to define a complete search space, i.e. the set of possible values each of the hyper-parameters can take. Unfortunately, in practical use cases this is often impossible due to insufficient knowledge about the problem at hand. Setting the hyper-parameter optimization bounds remains one of the main obstacles preventing wider use of the state-of-the-art HPO techniques \cite{pmlr-v51-shahriari16}.

HPO often needs to take into account specific requirements imposed by the environment where the model will run. This is exemplified by Marculescu et al. \cite{10.1145/3240765.3243479} in the context of hardware-aware ML. In order to deploy models to embedded and mobile devices, one needs to be aware of energy and memory constraints imposed by such devices. This creates a need for customized hardware-aware optimization techniques \change{that efficiently optimize for the accuracy of the model and the hardware jointly}.

\section{Model Verification}\label{sec:verification}
Verification is considered an essential step in any software development cycle as it ensures the quality of the product and reduces maintenance costs. As is the case with any software, ML models should generalize well to unseen inputs, demonstrate reasonable handling of edge cases and overall robustness, as well as satisfy all functional requirements. In this section, we discuss issues concerning three steps within model verification: requirement encoding, formal verification and test-based verification. 

\subsection{Requirement encoding}\label{sec:requirement}
Defining requirements for a machine learning model is a crucial prerequisite of testing activities. It often turns out that an increase in model performance does not translate into a gain in business value, as Booking.com discovered after deploying 150 models into production \cite{bernardi2019150}. One particular reason they highlight is \change{a} failure of proxy metrics (e.g. clicks) to convert to \change{the} desired business metric (e.g. conversion). Therefore\change{,} alongside accuracy measures, additional domain specific metrics need to be defined and measured. Depending on the application these may be inspired by KPIs and other business driven measures. In the case of Booking.com such metrics included conversion, customer service tickets or cancellations. \change{A} cross-disciplinary effort is needed to even define such metrics, as understanding from modeling, engineering and business angles is required. Once defined, these metrics should also be used for monitoring the production environment and for quality control of model updates.

Besides, simply measuring the accuracy of the ML model is not enough to understand its performance. Performance metrics should also reflect audience priorities. For instance Sato et al. \cite{sato2019cd} recommend validating models for bias and fairness, while in the case described by Wagstaff et al. \cite{10.1145/3292500.3330656} controlling for consumption of spacecraft resources is crucial.

\subsection{Formal Verification}\label{sec:formal-verification}
The formal verification step verifies that software functionality follows the requirements defined within the scope of the project. For ML models such verification could include mathematical proofs of correctness or numerical estimates of output error bounds, but as Ashmore et. al. \cite{ashmore2021assuring} point out this rarely happens in practice. More often\change{,} quality standards are being formally set via extensive regulatory frameworks that define what quality means and how models can be shown to meet them.

An example of where ML solutions have to adhere to regulations is the banking industry \cite{ananth2019opening}. This requirement was developed in the aftermath of the global financial crisis, as the industry realized that there was a need for heightened scrutiny towards models. As a consequence an increased level of regulatory control is now being applied to the processes that define how the models are built, approved and maintained. For instance, official guidelines \change{have} been published by the UK's Prudential Regulation Authority \cite{authority2018model} and European Central Bank \cite{guide2017guide}. These guidelines require \change{ML} model risk frameworks to be in place for all business decision-making solutions, and implementation of such frameworks requires developers to have extensive tests suites in order to understand \change{the} behavior of their ML models. The formal verification step in that context means ensuring that the model meets all criteria set by the corresponding regulations.

Regulatory frameworks share similarities with country-wide policies for governing \change{the} use of ML-powered technology, which we discuss in greater \change{detail} in Section~\ref{sec:ethics}.

\subsection{Test-based Verification}
In the context of ML, test-based verification is intended for ensuring that the model generalizes well to previously unseen data. While collecting a validation dataset is usually not a problem, as it can be derived from splitting the training dataset, it may not be \change{sufficient} for production deployment.

In an ideal \change{setting,} testing is done in a real-life setting, where business driven metrics can be observed, as we discussed in Section~\ref{sec:requirement}. Full scale testing in \change{a} real-world environment can be challenging for a variety of safety, security and scale reasons, and is often substituted with testing in simulation. That is the case for models for autonomous vehicles control \cite{kuutti2019survey}. Simulations are cheaper, faster to run, and provide flexibility to create situations rarely encountered in real life. Thanks to these advantages, simulations are becoming prevalent in this field. However, it is important to remember that simulation-based testing hinges on assumptions made by simulation developers, and therefore cannot be considered a full replacement for real-world testing. Even small variations between simulation and real world can have drastic effects on the system behavior, and therefore the authors conclude that validation of the model and simulation environment alone is not enough for autonomous vehicles. This point is emphasized further by the experiences from the field of reinforcement learning \cite{dulacarnold2019challenges}, where use of simulations is a de-facto standard for training agents.

Hackett et al. presented an instructive use case of how limited simulation-base\change{d} testing can be  \cite{hackett2018implementation}. The authors were \change{part} of \change{a} team that conducted an experiment that explored \change{a} reinforcement learning based cognitive engine (CE) for running a software-defined radio unit on board of the International Space Station (ISS). Preparation for the experiment included extensive ground testing in \change{an} emulated environment that informed many hyper-parameter choices and the computational setup. Nevertheless when the software was deployed on ISS, the actual conditions of the testing environment were so harsh the team was able to test only a subset of all planned experiments. The authors observed that despite extensive preparation CE was unable to cope with these emergency scenarios.

In addition, the dataset itself also needs to be constantly validated to ensure data errors do not creep into the pipeline and do not affect the overall quality. Data issues that go unnoticed can cause problems down the line that are difficult to troubleshoot. Breck et al. \cite{47967} argue that such issues are common in the setup where data generation is decoupled from the ML pipeline. Data issues can originate from bugs in code, feedback loops, changes in data dependencies. They can propagate and manifest themselves at different stages of the pipeline, therefore it is imperative to catch them early by including data validation routines \change{in} the ML pipeline.

\section{Model Deployment}
Machine learning systems running in production are complex software systems that have to be maintained over time. This presents developers with another set of challenges, some of which are shared with running regular software services, and some are unique to ML.

There is a separate discipline in engineering, called DevOps, that focuses on techniques and tools required to successfully maintain and support existing production systems. Consequently, there is a necessity to apply DevOps principles to ML systems. However, even though some of the DevOps principles apply directly, there are also a number of challenges unique to productionizing  machine learning. This is discussed in detail by Dang et al. \cite{dang2019aiops} which uses the term AIOps\footnote{Readers might have also encountered term MLOps (\url{https://ml-ops.org/}).} for DevOps tasks for ML systems. Some of the challenges mentioned include lack of high quality telemetry data as well as no standard way to collect it, difficulty in acquiring labels which makes supervised learning approaches inapplicable\footnote{Please refer to Section~\ref{sec:augmentation} for detailed discussion about data labeling.} and lack of agreed best practices around handling of machine learning models. In this section, we discuss issues concerning three steps within model deployment: integration, monitoring and updating. 

\subsection{Integration}\label{sec:integration}
The model integration step constitutes of two main activities: building the infrastructure to run the model and implementing the model itself in a form that can be consumed and supported. While the former is a topic that belongs almost entirely in systems engineering and therefore lies out of scope of this work, the latter is of interest for our study, as it exposes important aspects at the intersection of ML and software engineering. In fact, many concepts that are routinely used in software engineering are now being reinvented in the ML context.

Code reuse is a common topic in software engineering, and ML can benefit from adopting the same mindset. Reuse of data and models can directly translate into savings in terms of time, effort or infrastructure. An illustrative case is \change{an} approach Pinterest took towards learning image embeddings \cite{zhai2019learning}. There are three models used in Pinterest internally which use similar embeddings, and initially they were maintained completely separately, in order to make it possible to iterate on the models individually. However, this created engineering challenges, as every effort in working with these embeddings had to be multiplied by three. Therefore the team decided to investigate the possibility of learning a universal set of embeddings. It turned out to be possible, and this reuse ended up simplifying their deployment pipelines as well as improving performance on individual tasks.

A broad selection of engineering problems that machine learning practitioners now face is given in Sculley et al. \cite{sculley2015hidden}. \change{Many of them are known} anti-patterns in engineering\footnote{In software engineering an anti-pattern is understood as a common response to a recurring problem that is considered ineffective or counterproductive.}, but are currently widespread in machine learning software. Some of these issues, such as abstraction boundary erosion and correction cascades, are caused by the fact that ML is used in cases where the software has to take an explicit dependency on external data. Others, such as glue code or pipeline jungles, stem from the general tendency in the field to develop general-purpose software packages. Yet another source of problems discussed in the paper is the configuration debt: in addition to all configurations a regular software system may require ML systems \change{to} add a sizable number of ML-specific configuration settings that have to be set and maintained.

Researchers and software engineers often find themselves working together on the same project aiming to reach a business goal with a machine learning approach. On \change{the} surface there seems to be a clear separation of responsibilities: researchers produce the model while engineers build \change{the} infrastructure to run it. In reality, their areas of concern often overlap when considering the development process, model inputs and outputs and performance metrics. Contributors in both roles often work on the same code. Thus it is beneficial to include researchers in the whole development journey, making sure they own the product \change{codebase} along with the engineers, use the same version control and participate in code reviews. Despite obvious onboarding and slow-start challenges, this approach was seen to bring long term benefits in terms of speed and quality of product delivery \cite{amershi2019software}.

\subsection{Monitoring}\label{sec:monitoring}
Monitoring is one of the issues associated with maintaining machine learning systems as reported by Sculley et al. \cite{sculley2015hidden}. While monitoring is crucial for \change{the} maintenance of any software service, the ML community is in the early stages of understanding what are the key metrics of data and models to monitor and how to trigger system alarms when they deviate from normal behavior. Monitoring of evolving input data, prediction bias and overall performance of ML models is an open problem. Another maintenance issue highlighted by this paper that is specific to data-driven decision making is feedback loops. ML models in production can influence their own behavior over time via regular retraining. While making sure the model stays up to date, it is possible to create \change{a} feedback loop where the input to the model is being adjusted to influence its behavior. This can be done intentionally, as well as inadvertently, which is a unique challenge when running live ML systems.

Klaise et al. \cite{klaise2020monitoring} point out the importance of outlier detection as a key instrument to flag model predictions that cannot be used in a production setting. The authors name two reasons for such predictions to occur: the inability of the models to generalize outside of the training dataset and overconfident predictions on out-of-distribution instances due to poor calibration. Deployment of the outlier detector can be a challenge in its own right, because labeled outlier data is scarce, and the detector training often becomes a semi-supervised or even an unsupervised problem.

Additional insight on monitoring of ML systems can be found in Ackermann et al. \cite{ackermann2018deploying}. This paper describes an early intervention system (EIS) for two police departments in the US. On the surface their monitoring objectives seem completely standard: data integrity checks, anomaly detection and performance metrics. One would expect to be able to use out-of-the-box tooling for these tasks. However, the authors explain that they had to build all these checks from scratch in order to maintain good model performance. For instance, the data integrity check meant verifying updates of a certain input table and checksums on historical records, the performance metric was defined in terms of the number of changes in top \textit{k} outputs, and anomalies were tracked on rank-order correlations over time. All of these monitoring tools required considerable investigation and implementation. This experience report highlights a common problem with currently available end-to-end ML platforms: the final ML solutions are usually so sensitive to a problem's specifics that out-of-the-box tooling does not fit their needs well.

As a final remark we note that there is an overlap between \change{the} choice of metrics for monitoring and validation. The latter topic is discussed in Section~\ref{sec:requirement}.

\subsection{Updating}\label{sec:updating}
Once the initial deployment of the model is completed, it is often necessary to be able to update the model later on in order to make sure it always reflects the most recent trends in data and the environment. There are multiple techniques for adapting models to \change{new} data, including scheduled regular retraining and continual learning \cite{diethe2019continual}. Nevertheless in \change{the} production setting model updating is also affected by practical considerations.

A particularly important problem that directly impacts the quality and frequency of model update procedure is the concept drift, also known as dataset shift \cite{quinonero2009dataset}. Concept drift in ML is understood as changes observed in joint distribution $p(X, y)$, where $X$ is the model input and $y$ is the model output. Such changes can occur \change{discretely}, for example after some outside event that affects the input data distribution, or continuously, when data is gradually changing over time. Undetected, this phenomenon can have major adverse effects on model performance, as is shown by Jameel et al. \cite{jameel2020critical} for classification problems or by Celik and Vanschoren \cite{celik2021adaptation} in the AutoML context. Concept drift can arise due to a wide variety of reasons. For example, the finance industry faced turbulent changes as the financial crisis of 2008 was unfolding, and if advanced detection techniques were employed \change{they} could have provided additional insights into the ongoing crisis, as explained by Masegosa et al. \cite{masegosa2020analyzing}. Changes in data can also be caused by \change{an} inability to avoid fluctuations in the data collection procedure, as described in \change{the} paper \change{by} Langenk{\"a}mper et al. \cite{langenkamper2020gear} which studies the effects of slight changes in marine images on deep learning models' performance. Data shifts can have noticeable consequences even when occurring at \change{a} microscopic scale, as Zenisek et al. \cite{zenisek2019machine} show in their research on predictive maintenance for wear and tear of industrial machinery. Even though concept drift has been known for decades \cite{schlimmer1986incremental}, these examples show that it remains a critical problem for applications of ML today. \change{Consequently, detection of concept drift becomes a growing concern for teams that maintain ML models in production \cite{soemers2018adapting, sun2020review}, which makes this challenge directly related to monitoring discussed in the previous section.}

On top of the question of when to retrain the model to keep it up to date, there is an infrastructural question on how to deliver the model artifact to the production environment. In software engineering such tasks are commonly solved with continuous delivery (CD), which is an approach for accelerating \change{the} development cycle by building an automated pipeline for building, testing and deploying software changes. CD for machine learning solutions is complicated because, unlike in regular software products where changes only happen in the code, ML solutions experience change along three axes: the code, the model and the data. An attempt to formulate CD for ML as a separate discipline can be seen in Sato et al. \cite{sato2019cd}. This work describes the pieces involved and the tools that can be used at each step of building the full pipeline. A direct illustration of \change{the} benefits that a full CD pipeline can bring to the real-life ML solution can be found in Wider and Deger \cite{wider2019smart}.

While updating is necessary for keeping a model up to date with recent fluctuations in the data, it may also inflict damage on users or downstream systems because of the changes in \change{the} model's behavior, even without causing obvious software errors. To study updates in teams where AI is used to support human decision making, Bansal et al. \cite{bansal2019updates} introduced the notion of compatibility of an AI update. Authors define an update as compatible only if it does not violate \change{the} user's trust characterized via model behavior. \change{The} authors proceeded to show that updating an AI model to increase accuracy, at the expense of compatibility, may degrade overall AI-Human team performance. This line of research is continued by Srivastava et al. \cite{srivastava2020empirical}, who provide a detailed empirical study of backward compatibility issues in ML systems. They show how ML model updates may become backward incompatible due to optimization stochasticity or noisy training datasets. These findings motivate the need for de-noising and compatibility-aware training methods as the means to ensure reliable updates of deployed ML models.

\section{Cross-cutting aspects}
In this section we describe three additional aspects that ML projects have to consider: ethics, law, end users' trust, security. These aspects can affect every stage of the deployment pipeline.

\subsection{Ethics}\label{sec:ethics}
Ethical considerations should always inform data collection and modeling activities\footnote{Here by ``ethics'' we understand moral principles and techniques that inform the responsible development and use of ML or AI solutions.}. As stated in the report on ethical AI produced by the Alan Turing Institute \cite{leslie2019understanding}, ``it is essential to establish a continuous chain of human responsibility across the whole AI project delivery workflow''. If researchers and developers do not follow this recommendation, complications may come up due to a variety of reasons, some of which we illustrate in this section.

Since ML models use previously seen data to make decisions, they can rely on hidden biases that already exist in data - a behavior that is hard to foresee and detect. This effect is discussed in detail by O'Neil \cite{o2016weapons} in the field of criminal justice. Models that calculate \change{a} person's criminal “risk score” are often marketed as a way to remove human bias. Nevertheless, they use seemingly neutral demographic information like \change{a} neighborhood that often ends up serving as a proxy for more sensitive data such as race. As a result, people are disadvantaged on the basis of race or income. Machine translation is another example of an area where such hidden biases exist in data and can be exploited via an ML system. Prates et al. \cite{prates2020assessing} show \change{a} strong tendency towards male defaults in popular online translation service in particular for fields typically associated with unbalanced gender distribution, such as STEM.

Likewise, Soden et al. \cite{soden2019taking} mention \change{the} aggravation of social inequalities through \change{the} use of biased training datasets as one of the main hurdles in applying ML to Disaster Risk Management (DRM). It is argued that ML causes privacy and security\footnote{We discuss related cross-cutting security concerns in Section~\ref{sec:security}.} concerns through \change{a} combination of previously distinct datasets. Reducing \change{the} role of both experts and \change{the} general public is also seen as an ethical issue by DRM professionals, because they feel it increases \change{the} probability of error or misuse. Similar worries about unintentional or malign misuse of ML decision making systems are expressed by Muthiah et al. \cite{muthiah2016embers}. Their software for predicting civil unrest, called EMBERS, is designed to be used as a forecasting and communication tool, however authors remark that it can also be potentially misused by governments, either due to \change{a} misunderstanding of its role in society, or deliberately.

ML models for facial analysis often become subjects to criticism due to their unethical behavior. For example, Buolamwini and Gebru \cite{buolamwini2018gender} analyzed popular datasets for facial analysis and discovered them to be imbalanced on the basis of skin colour. They have also evaluated 3 commercial classification systems and showed darker-skinned females to be the most misclassified group. Authors conclude that urgent attention towards gender and skin type is needed for businesses that want to build genuinely fair facial analysis algorithms. Such questions about fairness and accountability of ML algorithms and data are studied by the branch of machine learning known as Fairness in ML \cite{barocas2017fairness}.

An interesting ethical aspect arises in \change{the} usage of ML in the field of creative arts, discussed by Anantrasirichai and Bull \cite{anantrasirichai2020artificial}. When a trained model is used to create a piece of visual art, it is not entirely clear where the authorship of this piece resides. The questions of originality therefore requires \change{special} attention. Closely related \change{to} this question is the growing concern of fake content being generated with ML, such as deepfake images and video, which can be easily used for the wrong purposes \cite{mirsky2020creation}.

\subsection{Law}\label{sec:law}
As ML grows its influence on society's everyday life, it is natural to expect more regulations to govern how ML models should function and how businesses, governments and other bodies can use them. Such legal frameworks can sometimes be used to guide decisions on ethics, although in general ethics and legal should be considered separate aspects.

Various countries have produced regulations to protect personal data rights. Typically, the more sensitive the information collected from the individual, the stronger the regulations governing its use. Examples of such regulations include the General Data Protection Regulation in \change{the} European Union \cite{rumbold2017effect} and ethical screening laws in a range of Asian countries \cite{aljunid2012health}. One domain that deals with some of the most sensitive information is healthcare. According to Han et al. \cite{han2020bridging}, many countries have strict laws in place to protect \change{the} data of patients, which makes \change{the} adoption of ML in healthcare particularly difficult. On one hand there is no doubt that these rules are absolutely necessary to make sure people are comfortable with their data being used. On the other hand\change{,} the amount of reviews, software updates and cycles of data collection/annotation that are required make it exceptionally hard to keep up with technical advances in ML, as Han et al. \cite{han2020bridging} explain following their experience deploying ML solutions in \change{the} healthcare sector in Japan.

Legislation takes time to develop, and often cannot keep up with the speed of progress in ML. This phenomenon is well known in \change{policymaking} and has been discussed in the context of ML as well as other technological advances by Marchant \cite{marchant2011growing} or more recently by \change{the} World Economic Forum \cite{malan2016values, malan2018law}. As Malan \cite{malan2018law} explains, by the time regulations are written they can already be out of date, resulting in a cat-and-mouse game that is wasteful on resources and causes legal framework abuses. Additionally, it is generally challenging to formulate specific and unambiguous laws for such a rapidly developing area as ML. For example, Wachter et al. show that GDPR lacks precise language as well as explicit and well-defined rights and safeguards, therefore failing to guarantee \change{the} ‘right to explanation’ \cite{wachter2017right}. As a result of these challenges, ML applications often have to abide by the existing laws of the area where they are deployed. Minssen et al. analyze the current regulatory approaches to medical ML in the US and Europe, and discuss how existing laws evaluate ML applications \cite{minssen2020regulatory}. At the same time governments have to provide ML adoption plans. For instance, the US Food and Drug Administration released an action plan outlining steps the agency plans to take to produce a regulatory framework for medical ML solutions \cite{stephens2021fda}. 

Companies should not be focusing solely on the technological side of their solutions, as DeepMind and Royal Free NHS Foundation Trust discovered while working on Streams, an application for automatic review of test results for serious conditions. Their initial collaboration was not specific enough on the use of patient data and on patient involvement overall, which triggered an investigation on their compliance with data protection regulations. The revised collaboration agreement was far more comprehensive and included \change{a} patient and public engagement strategy in order to ensure data is being used ethically \cite{suleyman2017information}.

\subsection{End users' trust}\label{sec:end-user}
ML is often met cautiously by the end users \cite{lai2020perceptions}, \cite{royal2017machine}, \cite{232961}. On their own accord\change{,} models provide minimal explanations, which makes it difficult to persuade end users of their utility \cite{han2020bridging}. In order to convince users to trust ML based solutions, \change{the} time has to be invested to build that trust. In this section, we explore ways in which that is done in practice.

If an application has a well-defined accessible audience, getting that audience involved early in the project is an efficient way to foster their confidence in the end product. This approach is very common in medicine, because the end product is often targeted at a well defined group of healthcare workers and/or patients. One example is the project called Sepsis Watch \cite{10.1145/3351095.3372827}. In this project the goal was to build a model that estimates \change{a} patient's risk of developing sepsis. It was not the first attempt at automating this prediction, and since previous attempts were considered failures, medical personnel were skeptical about \change{the} eventual success of Sepsis Watch. To overcome this skepticism, the team prioritized building trust, with strong communication channels, early engagement of stakeholders, front-line clinicians and decision makers, and established accountability mechanisms. One of the key messages of this work is that model interpretability has limits as a trust-building tool, and other ways to achieve high credibility with the end users should be considered. This aligns with conclusions made by ``Project explAIn'' which found that \change{the} relative importance of explanations of AI decisions varies by context \cite{ico2019explain}. A similar argument is made by Soden et al. \cite{soden2019taking}, who explore the impact ML has on disaster risk management (DRM). Due to \change{the} growing complexity of the ML solutions deployed, it is becoming harder for the public to participate and consequently to trust the ML-based DRM services, such as flooding area estimates or prediction of damage from a hurricane. As a mitigation measure the authors recommend making \change{the} development of these solutions as transparent as possible, by taking into account \change{the} voice of residents in the areas portrayed by models as ``at risk'' and relying on open software and data whenever possible. \change{The} importance of strong communication and engagement with early adopters is also emphasized by Mutembesa et al. \cite{mutembesa2018crowdsourcing} as they analyzed their experience of deploying a nation-wide cassava disease surveillance system in Uganda.

While the projects described above focused on engagement and accountability, in other circumstances explainability is the key to building \change{the} trust of the target audience. \change{This} is often the case when the users have experience and an understanding of ML. Rudin \cite{rudin2019stop} called the ML community to stop using black box models and explaining their behavior afterward, and instead design models that are inherently interpretable. Bhatt et al. \cite{bhatt2019explainable} analyzed explainability as a feature of machine learning models deployed within enterprises, and found that it is a must-have requirement for most stakeholders, including executives, ML engineers, regulators, and others. Moreover, their survey showed that explainability score is a desired model metric, along with measures of fairness and robustness. Explainability is also necessary in cases where it is demanded by the existing regulations\footnote{Effects of regulations on ML deployment are also discussed in Section~\ref{sec:formal-verification}}, and users will not trust decisions that are made automatically without provided explanations. Wang et al. \cite{wang2020using} describe such \change{a} requirement in the context of credit risk scoring. They observed that \change{the} XGBoost algorithm outperforms traditional scorecard approaches, but lacks \change{the} necessary explainability component. This prompted \change{the} authors to develop \change{a} custom loan decision explanation technique for XGBoost, subsequently deployed by QuickBooks Capital.

\change{A} poorly designed user interface can be one of the main obstacles in \change{the} adoption of any new technology. While this problem is not specific to ML, it is nevertheless worth mentioning it as a challenge ML applications face. For example, Wang et. al. studied \change{the} deployment of \change{the} AI-powered medical diagnosis tool "Brilliant Doctor" in rural China and discovered that the majority of doctors could not use it productively. One of the main reasons quoted was the UX design that did not take into account particularities of the environment (screen sizes, interaction with other software in the clinic) where it was installed, often resulting in an unusable software \cite{wang2021brilliant}. On the contrary, investing time in specialised user interfaces with tailored user experience can pay off with \change{fast} user adoption. Developers of Firebird \cite{madaio2016firebird}, a system that helps identify priority targets for fire inspection in the city of Atlanta, USA, found that the best way to avoid resistance from the end users while transitioning to an ML solution as a replacement of the previously used pen-and-paper method was to develop a user interface that presented the results of modelling in a way that the end users (fire officers and inspectors in the Fire dept) found most useful and clear. Similarly, authors of EMBERS \cite{muthiah2016embers}, a system that forecasts population-level events (such as protest), in Latin America, noticed that their users have two modes of using the system: (a) high recall: obtain most events, and then filter them using other methods; (b) high precision: focus on \change{a} specific area or \change{a} specific hypothesis. To improve the user experience and thus increase their confidence in the product, the user interface was improved to easily support both modes. This case study emphasizes the importance of context-aware personalization for ML systems' interfaces, one of the key observations delivered by ``Project explAIn'' \cite{ico2019explain}.

\subsection{Security}\label{sec:security}

Machine Learning opens up opportunities for new types of security attacks across the whole ML deployment workflow  \cite{kumar2019failure}. Specialised adversarial attacks for ML can occur on the model itself, the data used for training and also the resulting predictions. The field of adversarial machine learning studies the effect of such attacks against ML models and how to protect against them \cite{biggio2018wild, kurakin2016adversarial}. Recent work from Siva et al. found that industry practitioners are not equipped to protect, detect and respond to attacks on their ML systems \cite{siva2020adversarial}. In this section, we describe the three most common attacks reported in practice which \change{affect} deployed ML models: data poisoning, model stealing and model inversion. We focus specifically on adversarial machine learning and consider other related general security concerns in deploying systems such as access control and code vulnerabilities beyond the scope of our work.

In data poisoning, the goal of the adversarial attack is to deliberately corrupt the integrity of the model during the training phase in order to manipulate the produced results. In data poisoning scenarios an attacker is usually assumed to have access to the data that will ultimately be used for training, for example by sending emails to victims to subvert their spam filter \cite{nelson2008exploiting}. Poisoning attacks are particularly relevant in situations where the machine learning model is continuously updated with new incoming training data. Jagielski et al. reported that in a medical setting using a linear model, the introduction of specific malicious samples with \change{an} 8\% poisoning rate in the training set resulted in incorrect dosage for half of the patients~\cite{jagielski2018manipulating}.

Data poisoning can also occur as a result of a coordinated collective effort that exploits feedback loops we have discussed in Section~\ref{sec:monitoring}, as happened with Microsoft's Twitter bot Tay \cite{schwartz20192016}. Tay was designed to improve its understanding of the language over time but was quickly inundated with a large number of deliberately malevolent tweets. Within 16 hours of its release a troubling percentage of Tay's messages were abusive or offensive, and the bot was taken down.

Another type of adversarial attack is reverse engineering a deployed model by querying its inputs (e.g. via a public prediction API) and monitoring the outputs. The adversarial queries are crafted to maximize the extraction of information about the model in order to train a substitute model. This type of attack is referred to as model stealing. In a nutshell, this attack results in \change{the} loss of intellectual property which could be a key business advantage for the defender. Tramèr et al. \cite{tramer2016stealing} have shown that it is possible to replicate models deployed in production from ML services offered by Google, Amazon and Microsoft across a range of ML algorithms including logistic regression, decision trees, SVMs and neural networks. In their work, they report the number of queries ranging from 650 to 4013 to extract an equivalent model and in time ranging from 70s to 2088s.

A related attack is that of model inversion where the goal of the adversarial attack is to recover parts of the private training set, thereby breaking its \change{confidentiality}. Fredrikson et al. have shown that they could recover training data by exploiting models that report confidence values along \change{with} their predictions \cite{fredrikson2015model}. Veale et al. \cite{veale2018algorithms} emphasize the importance of protecting against model inversion attacks as a critical step towards compliance with data protection laws such as GDPR.

\section{Discussion of potential solutions}
This survey looked at case studies from a variety of industries: computer networks, manufacturing, space exploration, law enforcement, banking, and more. However, further growth of ML adoption can be severely hindered by poor deployment experience. To make the ML deployment scalable and accessible to every business that may benefit from it, it is important to understand the most critical pain points and to provide tools, services and best practices that address those points. We see this survey as an initial step in this direction: by recognizing the most common challenges currently being reported we hope to foster an active discussion within the academic community about what possible solutions might be. We classify possible research avenues for solutions into two categories, which we discuss below. We also give some concrete examples, but since the purpose of this section is illustrative, we do not aim to provide a complete survey of ML tools and development approaches.

\subsection{Tools and services}\label{sec:tools}
The market for machine learning tools and services is experiencing rapid growth \cite{huyen2020tools}. As a result, tools for individual deployment problems are continuously developed and released. Consequently, some of the problems we have highlighted can be solved with the right tool.

For example, this is most likely the case for operational maintenance of ML models, discussed in Sections~\ref{sec:monitoring} and \ref{sec:updating}. Many platforms on the market offer end-to-end experience for the user, taking care of such things as data storage, retraining and deployment. Examples include AWS SageMaker \cite{venkateswar2019using}, Microsoft ML \cite{team2016azureml}, Uber Michelangelo \cite{hermann2017meet}, TensorFlow TFX \cite{baylor2019continuous}, MLflow~\cite{zaharia2018accelerating} and more. A typical ML platform would include, among other features, \change{a} data storage facility, model hosting with APIs for training and inference operations, a set of common metrics to monitor model health and an interface to accept custom changes from the user. By offering managed infrastructure and a range of out-of-the-box implementations for common tasks such platforms greatly reduce \change{the}  operational burden associated with maintaining the ML model in production.

Quality assurance, which is the focus of Section~\ref{sec:verification}, also looks to be an area where better tools can be of much assistance. Models can greatly benefit from \change{the} development of a test suite to verify their behavior, and the community actively develops tools for that purpose. Jenga \cite{schelter2021jenga} ensures model's robustness against errors in data, which very commonly occur in practice as was mentioned in Section \ref{sec:data}. CheckList methodology \cite{acl20checklist} provides a formal approach towards assessing \change{the} quality of NLP models. The Data Linter \cite{hynes2017data} inspects ML data sets to identify potential issues in the data.

As discussed in Section~\ref{sec:augmentation}, obtaining labels is often a problem with real world data. Weak supervision has emerged as a separate field of ML which looks for ways to address this challenge. Consequently, a number of weak supervision libraries are now actively used within the community, and show promising results in industrial applications. Some of the most popular tools include Snorkel \cite{bach2019snorkel}, Snuba \cite{varma2018snuba} and cleanlab \cite{northcutt2019confidentlearning}.

A growing field of AutoML \cite{li2021automl} aims to address challenges around \change{a} model selection and hyper-parameter tuning, discussed in Sections \ref{sec:model-selection} and \ref{sec:hpo}. There is a large variety of tools that provide general-purpose implementations of AutoML algorithms, such as Auto-keras \cite{jin2019auto}, Auto-sklearn \cite{NIPS2015_11d0e628} or TPOT \cite{olson2016tpot}. However practical reports of applying AutoML to real world problems indicate that practitioners need to exercise extreme caution, as AutoML methods might not be ready for decision making in high-stakes areas yet \cite{ebadi2019can, faes2019automated}.

\change{Given the potential damage an unnoticed dataset shift can cause to the quality of predictions of deployed ML models (see Section~\ref{sec:updating}), there are many techniques to detect and mitigate its effects. A large variety of methods based on dimensionality reduction and statistical hypothesis testing is reviewed by Rabanser at al.~\cite{Rabanser2019FailingLA}, and are now implemented in software libraries (e.g. Alibi Detect~\cite{Van_Looveren_Alibi_Detect_Algorithms_2022}) and services (e.g. Azure ML\footnote{\url{https://docs.microsoft.com/en-us/azure/machine-learning/how-to-monitor-datasets}}and AWS Sagemaker\footnote{\url{https://docs.aws.amazon.com/sagemaker/latest/dg/clarify-model-monitor-feature-attribution-drift.html}}), available for use in deployment's monitoring suites. The community has also made strides in developing strategies for dealing with the shift once it was detected. We refer interested readers to the works on using domain adaptation \cite{gama2014survey}, meta-learning \cite{finn2017model} and transfer learning \cite{zhao2014online, xie2017selective} as ways of addressing dataset shift.}

Using specific tools for solving individual problems is a straightforward approach. However practitioners need to be aware that by using a particular tool they introduce an additional dependency into their solution. While a single additional dependency seems manageable, their number can quickly grow and become a maintenance burden. Besides, as we mentioned above, new tools for ML are being released constantly, thus presenting practitioners with the dilemma of choosing the right tool by learning its strengths and shortcomings.

\subsection{Holistic approaches}
Even though ML deployments require software development, ML projects are fundamentally different from traditional software engineering projects. The main differences arise from unique activities like data discovery, dataset preparation, model training, deployment success measurement, etc. \change{Some of these activities cannot be defined precisely enough to have a reliable time estimate (as discussed in Section~\ref{sec:data-collection} in regard to data collection), some require a different style of project management (Section~\ref{sec:integration} discusses the challenges of managing mixed team engineers and scientists), and some make it difficult to measure the overall added value of the project (see Section~\ref{sec:requirement} for the discussion of translation of ML model performance to the business value). For these reasons ML deployment projects often} do not lend themselves well to widespread approaches to software engineering management paradigms, and neither to common software architectural patterns \change{\cite{zujus2018ai}}.

Compared to classical software engineering (SE), ML introduces unique artifacts with unique characteristics: datasets and models. \change{Unlike program source and configuration files, these artifacts are not distributed as program code, and exist in the form of tabular or binary files. Regular SE tools for such common operations as source control, branching and merging, review, cannot be applied to these new artifacts "as is" \cite{barrak2021co}. Consequently it is essential to develop documentation approaches that are most suitable for these artifacts.} There is a growing body of literature that aims to adapt existing or develop new practices for handling datasets and models in a coherent and reliable way. Lawrence \cite{lawrence2017data} proposes an approach toward classifying \change{the} readiness of data for ML tasks, which Royal Society DELVE applied \change{in one} of their reports in the context of COVID-19 pandemic \cite{delve-7}. Gebru et al. suggested ``datasheets for datasets'' \cite{gebru2018datasheets} in order to document dataset's motivation, composition, collection process and intended purposes. \change{Data Version Control (DVC) is an open-source project that aims to create a Git-like source control experience for datasets \cite{ruslan_kuprieiev_2022_6195393}.} For models, Mitchell et al. \cite{mitchell2019model} proposed model cards, short documents that accompany trained models detailing their performance, intended use context, and other information that might be relevant for model's application.

Data Oriented Architectures (DOA, \cite{lawrence2019doa}, \cite{borchert2020milan}, \change{\cite{paleyes2021towards}}) is an example of an idea that suggests rethinking how things are normally approached in software development, and by doing so promises to solve many of the issues we have discussed in this survey. Specifically, the idea behind DOA is to consider replacing micro-service architecture, widespread in current enterprise systems, with dataflow-based architectures, thus making data flowing between elements of business logic more explicit and accessible. Micro-service architectures have been successful in supporting high scalability and embracing the single responsibility principle. However, they also make data flows hard to trace, and it is up to owners of every individual service to make sure inputs and outputs are being stored in a consistent form (these issues are also discussed Section~\ref{sec:data-collection}). DOA provides a solution to this problem by moving data to streams flowing between stateless execution nodes, thus making data available and traceable by design, therefore making simpler the tasks of data discovery, collection and labeling. In its essence DOA proposes to acknowledge that modern systems are often data-driven, and therefore need to prioritize data in their architectural principles.

As noted above, ML projects normally do not fit well with commonly used management processes, such as Scrum or Waterfall. Therefore it makes sense to consider processes tailored specifically for ML. One such attempt is done by Lavin et. al. \cite{lavin2021technology}, who propose Machine Learning Technology Readiness Levels (MLTRL) framework. MLTRL describes a process of producing robust ML systems that takes into account key differences between ML and traditional software engineering with \change{a} specific focus on \change{the} quality of \change{the} intermediate outcome of each stage of the project. As we discussed in Section~\ref{sec:verification}, verification is the area of ML deployment that suffers from \change{a} lack of standard practices, and in that context MLTRL suggests a possible way to define such standards.

A very widespread practice in software engineering is to define a set of guidelines and best practices to help developers make decisions at various stages of the development process. These guidelines can cover a wide range of questions, from variable names to execution environment setup. For example, Zinkevich \cite{zinkevich2017rules} compiled a collection of best practices for machine learning that are utilized in Google. While this cannot be viewed as a coherent paradigm \change{for} doing ML deployment, this document gives practical advice on a variety of important aspects that draw from the real life experiences of engineers and researchers in the company. Among others, rules and suggestions for such important deployment topics as monitoring (discussed in Section~\ref{sec:monitoring}), end user experience (Section~\ref{sec:end-user}) and infrastructure(Section~\ref{sec:integration}) are described.

Besides serving as a collection of advice for common problems, guidelines can also be used as a way to unify approaches towards deploying ML in \change{a} single area. The Association of German Engineers (VDI) has released a series of guidelines on various aspects of big data applications in \change{the} manufacturing industry \cite{vdi2019guidelines}. These documents cover a wide range of subjects, including data quality, modeling, user interfaces, and more. The series aims to harmonize the available technologies used in the industry, facilitate cooperation and implementation. Such initiatives can help bridge the gap between ML solutions and regulations in a particular applied area discussed in Section~\ref{sec:law} of this survey.

Holistic approaches are created with ML application in mind, and therefore they have the potential to offer \change{significant} ease of deploying ML. But it should be noted that all such approaches assume significant time investment, because they represent significant changes to current norms in project management and development. Therefore a careful assessment of risks versus benefits should be carried out before adopting any of them. 

\section{Further Work}
Even though the set of challenges we reviewed covers every stage of the ML deployment workflow, it is far from complete. Identifying other, especially non-technical, challenges is a natural direction of further work and could be augmented by conducting interviews with industry representatives about their experiences of deploying ML.

In this paper, we reviewed reports from a variety of industries, which shows the ubiquity and variety of challenges with deploying ML in production. An interesting extension would be the comparative analysis of industries. Quantitative and qualitative analysis of most commonly reported challenges may open interesting transferability opportunities, as approaches developed in one field may be applicable in the other.

Our work includes a brief discussion of existing tools in Section~\ref{sec:tools}. However, the community would benefit from a comprehensive review of currently available tools and services, mapped to challenges reported in our study. This new work could be combined with our survey to enable practitioners to identify the problem they are facing and choose the most appropriate tool that addresses that problem.

\section{Conclusion}
In this survey, we showed that practitioners deal with challenges at every step of the ML deployment workflow due to practical considerations of deploying ML in production. We discussed challenges that arise during the data management, model learning, model verification and model deployment stages, as well as considerations that affect the whole deployment pipeline including ethics, end users' trust, \change{law} and security. We illustrated each stage with examples across different fields and industries by reviewing case studies, experience reports and the academic literature. 

We argue that it is worth \change{the} academic community's time and focus to think about these problems, rather than expect each applied field to figure out their own approaches. We believe that ML researchers can drive improvements to the ML deployment experience by exploring holistic approaches and taking into account practical considerations.

ML shares a lot of similarities with traditional computer science disciplines, and consequently faces similar challenges, albeit with its own peculiarities. Therefore ML as a field would benefit from a cross-disciplinary dialog with such fields as software engineering, human-computer interaction, systems, \change{policymaking}. Many pain points we have described in this work were already experienced by communities in these fields, and the ML community should turn to them for solutions and inspiration.

As an observation that follows from the process of collecting papers to review in this survey, we note the \change{relative} shortage of deployment experience reports in the academic literature. Valuable knowledge obtained by industry ML practitioners goes unpublished. We would like to encourage organisations to prioritize sharing such reports, as they provide valuable information for the wider community, but also as a way to self-reflect, collect feedback and improve on their own solutions. \change{Nevertheless, several venues focused ML deployment already exist, and we encourage interested readers to follow them:
\begin{itemize}
	\item International Conference Knowledge Discovery and Data Mining (ACM SIGKDD), Applied Data Science track.
	\item Conference on Innovative Applications of Artificial Intelligence (IAAI).
	\item European Conference on Machine Learning and Principles and Practice of Knowledge Discovery in Databases (ECML PKDD), Applied Data Science track.
	\item IEEE International Conference on Machine Learning and Applications (ICMLA).
	\item Applied Artificial Intelligence journal.
	\item Neural Computing and Applications journal.
\end{itemize}
}

We hope this survey will encourage discussions within the academic community about pragmatic approaches to deploying ML in production.

\begin{acks}
A.P. and N.L. are grateful for funding from a Senior AI Fellowship from the Alan Turing Institute (ATI) and UK Research \& Innovation (UKRI). We would like to thank our reviewers for thoughtful and detailed comments that helped improve the paper. We would also like to thank Jessica Montgomery, Diana Robinson and Ferenc Huszar for insightful discussions.
\end{acks}

\bibliographystyle{unsrt}
\bibliography{references}

\end{document}